%% file: main.tex
\documentclass[10pt,twocolumn,letterpaper]{article}

\usepackage{wacv} 

\definecolor{wacvblue}{rgb}{0.21,0.49,0.74}
\usepackage{xcolor}

\DeclareMathOperator{\success}{success}

\input{preamble}
\usepackage[pagebackref,breaklinks,colorlinks,allcolors=wacvblue]{hyperref}
\usepackage{orcidlink}

\def\wacvPaperID{1052} % *** Enter the WACV Paper ID here
\def\confName{WACV}
\def\confYear{2027}

\makeatletter
\newcommand{\blfootnote}[1]{%
  \begingroup
  \renewcommand\thefootnote{}%
  \footnotetext{#1}%
  \addtocounter{footnote}{-1}%
  \endgroup
}
\makeatother

\newcommand{\SE}{SE\xspace}
\newcommand{\CE}{CE\xspace}
\newcommand{\NPS}{NSE\xspace}
\newcommand{\NPC}{NCE\xspace}
\newcommand{\GAN}{GAN\xspace}
\newcommand{\DM}{DM\xspace}

\title{Generating Medical Image Counterfactuals using Causal Explanations}

\author{
David A. Kelly$^{*,1}$\orcidlink{0000-0002-5368-6769},
Tom Yaacov$^{*,1}$\orcidlink{0000-0002-0565-6506},
Nathan Blake$^{1}$\orcidlink{0000-0002-6404-514X},
Sander Beckers$^{2}$\orcidlink{0000-0002-9202-0644},
and Hana Chockler$^{1}$\orcidlink{0000-0003-1219-0713}
}

\begin{document}

\maketitle 

\blfootnote{$^{1}$ King's College London,
{\tt\footnotesize \{firstname.lastname\}@kcl.ac.uk}}

\blfootnote{$^{2}$ University College London,
{\tt\footnotesize srekcebrednas@gmail.com}}

\blfootnote{* Equal contribution.}

\blfootnote{** This work was supported by the Causality in Healthcare AI (CHAI) Hub
[UKRI AI and EPSRC grant EP/Y028856/1].}

\begin{abstract}

Deep learning models have achieved impressive performance in medical image diagnosis, yet their deployment in clinical settings remains constrained by limited explainability. Counterfactual images provide one means of auditing model behavior by showing how an image would need to change for a classifier to produce a different prediction. Existing approaches typically generate such explanations using auxiliary models, including generative adversarial networks and diffusion models. While often capable of producing visually realistic images, these methods explain one black-box model using another, making it difficult to separate the classifier's decision-making process from the inductive biases of the generator.

We propose a novel counterfactual-generation framework that requires no generative model. Instead, counterfactuals are constructed directly from causal evidence extracted from the classifier. The resulting approach is deterministic, requires no additional model training, and enables controllable edits within user-specified regions of interest. Experiments on real-world medical imaging datasets demonstrate that the proposed method successfully changes classifier predictions while remaining closer to the original image than generative baselines, providing a more direct and transparent view of the classifier's decision boundary.

% Deep learning for medical imaging diagnostics faces the problem of a lack of explainability. Counterfactual images have been proposed as one method of providing explainability: for instance, taking an image the model classifies as healthy and altering it so it is classified as diseased. This gives healthcare practitioners access to `what-if' example images. Most existing methods for generating such counterfactuals rely on deep learning techniques, particularly generative models such as generative adversarial networks and diffusion models. While these approaches can be effective at generating counterfactuals, they inherit the same fundamental limitation as the models they aim to explain: they operate as unexplained black boxes. In this paper, we propose a novel approach for generating counterfactuals by using the framework of causal explanations. Our method performs targeted, local modifications to the input image by using causal explanations derived from images in the positive class, shifting the decision of the model in a controlled and interpretable manner. Contrary to existing work, our process is fully transparent and produces counterfactuals that remain close to the original image. We demonstrate the usefulness of our approach through evaluations on real-world medical datasets.

%\keywords{Counterfactual Reasoning \and Explainable AI \and Medical Imaging.}
% Authors must provide keywords and are not allowed to remove this Keyword section.

\end{abstract}

\section{Introduction}

Deep image classifiers achieve strong performance across many medical-imaging tasks, yet their use remains limited by poor generalization, dataset shift, and a lack of transparent explanations~\cite{blake2024explainable}. These concerns are not unique to healthcare: they reflect a broader challenge in computer vision, where high-capacity models may produce accurate predictions while relying on features that are difficult for domain experts to inspect. In medical imaging, the cost of such opacity is especially acute. A clinician or model auditor may need to know not only \emph{where} a classifier attends, but also \emph{what visual evidence would have changed its decision}.

Counterfactual explanations address this question by asking how an input image would need to change for a classifier to assign it to a different class~\cite{wachter2017counterfactual}. In medical imaging, this often takes the form of transforming an image classified as healthy into one classified as diseased. Such `what-if' images can help users probe the decision boundary of a model and identify its modes of failure. This differs from many post-hoc explainable AI (XAI) methods such as \plainshap~\cite{lundberg2017unified}, \lime~\cite{ribeiro2016should}, or \rex~\cite{CKKS26}, which identify those features responsible for a given classification. These attribution methods ask \emph{which} features support the current prediction; counterfactual generation asks \emph{what would need to change} for the prediction to change. 
This is not the same as simulating plausible pathological counterfactuals, but rather a \emph{model-specific} audit. A small non-pathological alteration may induce a class to change to 'disease': such a discrepancy is precisely what the method seeks to expose. 

Most existing approaches to image counterfactual generation rely on generative models, including GANs, conditional GANs, and diffusion models~\cite{mertes2022ganterfactual,singla2023explaining,liu2024controllable,oh2022learn,zigutyte_2024}. These methods synthesize plausible samples from a target class and have been widely adopted in medical imaging because they can model complex anatomical and pathological variation~\cite{singla2023explaining}. However, they introduce a fundamental challenge for explainability: the explanation of one black-box model is produced by another. As a result, it can be difficult to determine whether a counterfactual reflects the classifier's decision boundary, the generative model's image prior, or artifacts of the optimization procedure. Generative approaches are also typically data-intensive and sensitive to training choices, which is problematic in medical imaging where large, well-annotated datasets are often scarce~\cite{cohen2021gifsplanation}. For model auditing, the most informative counterfactual is not necessarily the most realistic target-class image, but a close image that the classifier itself accepts as belonging to the target class. 

In this paper, we introduce a non-generative approach to image counterfactual generation. Instead of learning a separate image-generation process, we directly exploit the classifier's own causal explanations (\Cref{subsec:causality}) to construct targeted counterfactual perturbations. We pre-compute causal explanations for positively classified images, retrieve explanations from images close to the target instance, and partially transfer the associated causal pixel values to generate candidate counterfactuals. Consequently, the search space is constrained entirely by evidence that the classifier has already identified as causally relevant, rather than by a learned image prior. 

\begin{figure}[t]
\centering
    \begin{subfigure}[t]{0.3\linewidth}
        \centering
        \includegraphics[scale=0.25]{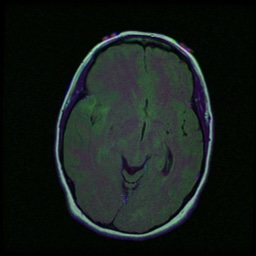}
        \caption{Brain MRI with negative\\classification.}\label{fig:false_neg}
    \end{subfigure}
%\hfill
    \begin{subfigure}[t]{0.3\linewidth}
        \centering
        \includegraphics[scale=0.25]{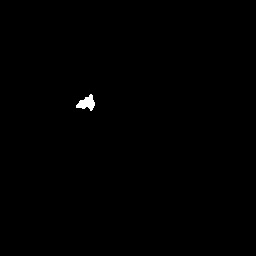}
        \caption{Human provided annotation.}\label{fig:tumor_loc}
    \end{subfigure}
%\hfill
% --- Column 2 ---
    \begin{subfigure}[t]{0.3\linewidth}
        \centering
        \includegraphics[scale=0.25]{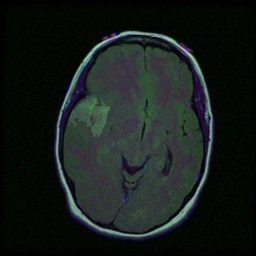}
        \caption{A minimal change sufficient for class\\`tumor'.}%
        \label{fig:tumor_pos}
    \end{subfigure}
\caption{A clinician may be surprised to find~\Cref{fig:false_neg} classed as `no tumor'. In fact, there is a tumor present -- ~\Cref{fig:tumor_loc} -- that the model has failed to detect. Our approach generates an approximately minimal counterfactual, where `minimal' stands for a smallest change to the smallest number of pixels.}%
\label{fig:false-negative}
\end{figure}

To our knowledge, this is the first counterfactual-generation framework that 
directly reuses causal explanation. Our method offers three advantages for visual model auditing. First, it requires no additional trained models, isolating the behavior of the classifier under inspection and eliminating generator-induced confounding effects. Second, the procedure is deterministic, avoiding the stochasticity common to training and sampling from generative models. Third, perturbations are local and controllable: users can restrict the search to a region of interest and directly test whether modifying a specific anatomical structure is sufficient to alter the classifier's prediction (\Cref{fig:false-negative}).

% We evaluate our method on brain MRI and skin-lesion datasets against GAN- and diffusion-based baselines. Across both domains, causal counterfactuals achieve successful prediction changes with smaller image modifications, demonstrating that classifier-derived causal explanations can probe the decision boundary with smaller changes than generative counterfactual models (\Cref{sec:results}).

Conceptually, our work reframes counterfactual generation as a causal-retrieval problem rather than a generative-modeling problem. Instead of learning how to synthesize target-class images, we identify and transfer causal evidence already used by the classifier when making positive predictions. This produces counterfactuals that are inherently model-specific: every modification can be traced to a previously observed causal explanation. As a result, the generated images provide a more direct view of the classifier's decision boundary and avoid the ambiguity introduced by auxiliary generative models.

Our principle contributions are:
\begin{itemize}
    \item The first counterfactual framework built directly from causal explanations. 
    \item Counterfactuals are constrained by classifier-derived causal evidence rather than image realism priors.
    \item Every perturbation is traceable to an identified causal contribution.
    \item Counterfactual generation becomes a model-auditing procedure rather than an image-synthesis procedure.
\end{itemize}

The complete code for the evaluation and the implementation are available in~\cite{supplementarymaterial}.

\section{Background}\label{sec:Background}

This section introduces the necessary background that is developed in greater detail in \cite{kelly2025causal}. The general idea is that we can interpret a classifier as a \emph{causal model} \citep{Pearl09a,Hal19}, where the (causally) independent pixels $\vec{X}$ jointly cause the output classification $O$ via the causal mechanism implemented by the classifier $\mathcal{N}$. Doing so allows us to invoke well-defined notions of causal explanations that are grounded in the theory of \emph{actual causality}~\citep{Hal19,beckers22}. We briefly review a simplified version of the relevant concepts and refer the reader to~\cite{kelly2025causal} for a comprehensive treatment. 

\subsection{Causal Explanations}\label{subsec:causality}
Let the set of variables $\vec{X}$ correspond to the set of pixels of an image, so that any setting $\vec{X}=\vec{x}$ corresponds to a specific image. Let $O$ be the output variable of a classifier $\mathcal{N}$, so that $\mathcal{N}(\vec{x})=o$ means the image $\vec{X}=\vec{x}$ is classified as belonging to the class $O=o$. 

In order to derive \emph{sufficient} explanations of the classification, we evaluate the behavior of the classifier on images that are the  result of placing a \emph{mask} onto the original image. For each $X_i \in \vec{X}$, we let the Boolean variable $V_i$ denote whether the corresponding pixel is masked -- $V_i = 1$ -- or takes on its original pixel value -- $V_i=0$. A mask is implemented using a predefined \emph{masking value}, or \emph{baseline value}.

Given a masking image $\vec{V}=\vec{v}$ and an original image $\vec{X}=\vec{x}$, placing the mask onto the original image results in the Hadamard product of $\vec{X}=\vec{x}$ and $\vec{V}=\vec{v}$, denoted as $\vec{v} \odot \vec{x}$. If the masked image does not alter the classification, then the unmasked part of the original image suffices for the classification. For it to also explain the classification, the unmasked part should not contain any redundant pixels, and thus we also demand that it is minimal. Concretely, we define sufficient explanations as follows.

\begin{definition}[Sufficient Explanation (MSPS)~\cite{CKKS26}]\label{defn:simple-exp}
Given an input image $\vec{X}=\vec{x}$ and a classifier 
$\mathcal{N}$, we say that a subset $\vec{X}_{exp}=\vec{x}_{exp}$ of $\vec{X}=\vec{x}$ is a \emph{sufficient explanation} of the classification $\mathcal{N}(\vec{x})$ 
if the following conditions hold:
\begin{description}
\item[{\rm EXIM1.}] $\mathcal{N}(\vec{v}_{-exp} \odot \vec{x})=\mathcal{N}(\vec{x})$.
\item[{\rm EXIM2.}] $\vec{X}_{exp}$ is minimal; there is no strict subset $\vec{X}'_{exp}$ of $\vec{X}_{exp}$ that satisfies EXIM1.
\end{description}
Here $\vec{V}=\vec{v}_{-exp}$ is the mask over all pixels not in $\vec{X}_{exp}$.
\end{definition}

In short, sufficient explanations are simply \emph{Minimal Sufficient Pixel Sets} -- MSPS.

\cite{kelly2025causal} also introduce the stronger notion of \emph{complete explanations} for image classifiers.
Informally, a complete explanation is an explanation that is both sufficient and necessary, and minimally so.

\dfn[Complete Explanation\label{def:complete}]
Given an input image $\vec{X}=\vec{x}$ and a classifier 
$\mathcal{N}$, we say that a subset $\vec{X}_{exp}=\vec{x}_{exp}$ of $\vec{X}=\vec{x}$ is a \emph{complete explanation} of the classification $\mathcal{N}(\vec{x})$ 
if it satisfies EXIM1 and the following conditions hold:
\begin{description}
    \item[\rm {EXN1.}] $\mathcal{N}(\vec{v}_{exp} \odot \vec{x}) \neq \mathcal{N}(\vec{x})$.
    \item[\rm {EXN2.}] $\vec{X}_{exp}$ is minimal; there is no strict subset of $\vec{X}_{exp}$ that satisfies both EXIM1 and EXN1.
\end{description}
\edfn

\subsection{ReX}
\label{sec:rex}

We use the tool \rex~\cite{CKKS26} to generate the causal explanations just described. \rex is an \xai tool based in the theory of actual causality~\citep{Hal19,beckers22}.
We use \rex because, unlike other \xai tools, can provide us with causal explanations directly. We provide here a brief overview of the algorithm and direct the interested reader to~\cite{CKKS26} for full details.

\rex starts by dividing an image into $4$ parts. We call each part a superpixel. \rex creates \emph{mutants} of the original image by covering all combinations of superpixels with a baseline value (typically $0$). 
These combinations are tested against the model and sorted between those which satisfy the required classification and those which do not. Causal responsibility is distributed over the different superpixels,
where responsibility is a quantitative measure of causality and, broadly speaking, measures the amount of causal influence on the classification~\citep{CH04,CKKS26}. Passing superpixel combinations are further refined into smaller superpixel combinations and tested using the same procedure.

In this way, mutant generation is iteratively guided by the model, and responsibility tends to concentrate on those pixels which occur most frequently in class preserving combinations. This is repeated many times, starting from a different random partition of the image. 
This procedure provides a ranking of pixels based on their responsibility.
The pixels are ranked in the order of their responsibility for the classification and a causal explanation, or a \emph{minimal sufficient pixel set} (MSPS) (Def. \ref{defn:simple-exp}), is extracted from this ranking using a greedy algorithm. The number of calls \rex makes to the model to calculate initial causal explanations is $O(2^s n N)$, where $s$ is the size of the partition in each step (by default $s=4$), $n$~is the number of pixels in the original image $x$, and $N$ is the number of initial partitions~\cite{CKKS26}. 
\rex is also capable of discovering multiple explanations~\cite{CKK25} but these do not occur in our 
datasets.

\rex also computes \emph{complete explanations} (Def. \ref{def:complete}), minimal sets of pixels which are both sufficient and necessary for the original classification~\cite{kelly2025causal}. Recall, that this means the pixels, by themselves, are sufficient against the baseline, and, when removed from the original image, they change the classification. These are, in general, larger than sufficient explanations and contain more information. We evaluate both sufficient and complete explanations.

\section{Causal Explanations and Counterfactual Generation}
\label{sec:method}

\definecolor{lightblue}{HTML}{00b0f0}

\begin{figure*}[t]
    \centering
    \begin{tikzpicture}[
    background rectangle/.style=
    {draw=red!30,fill=orange!10,rounded corners=1ex},
    show background rectangle,
    arrow/.style={
        -Latex, 
        color=lightblue,
        line width=1.3mm,
        }
        ]
        
        \node[minimum size=2cm,label={[align=center]above:\ding{192} negatively classified\\image (`no tumor')}] at (0,0) (start) {
            \includegraphics[scale=0.2]{images/TCGA_DU_6408_19860521_21/original.jpg}
        };
        \node[draw,fill=blue!10,minimum size=2cm, rounded corners=1ex] at (3,0) (counterfactual) {What if?};
        
        \node[minimum size=2cm,label={[align=center]above:\ding{193} nearby positive\\(diseased) images}] at (6,0) (nearby) {
            \includegraphics[scale=0.15]{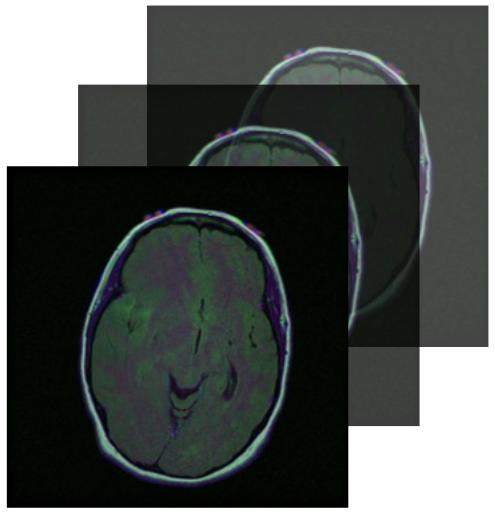}
        };

        \node[minimum size=2cm,label={[align=center]above:\ding{194} pre-computed\\explanations}] at (9,0) (extract) {
            \includegraphics[scale=0.13]{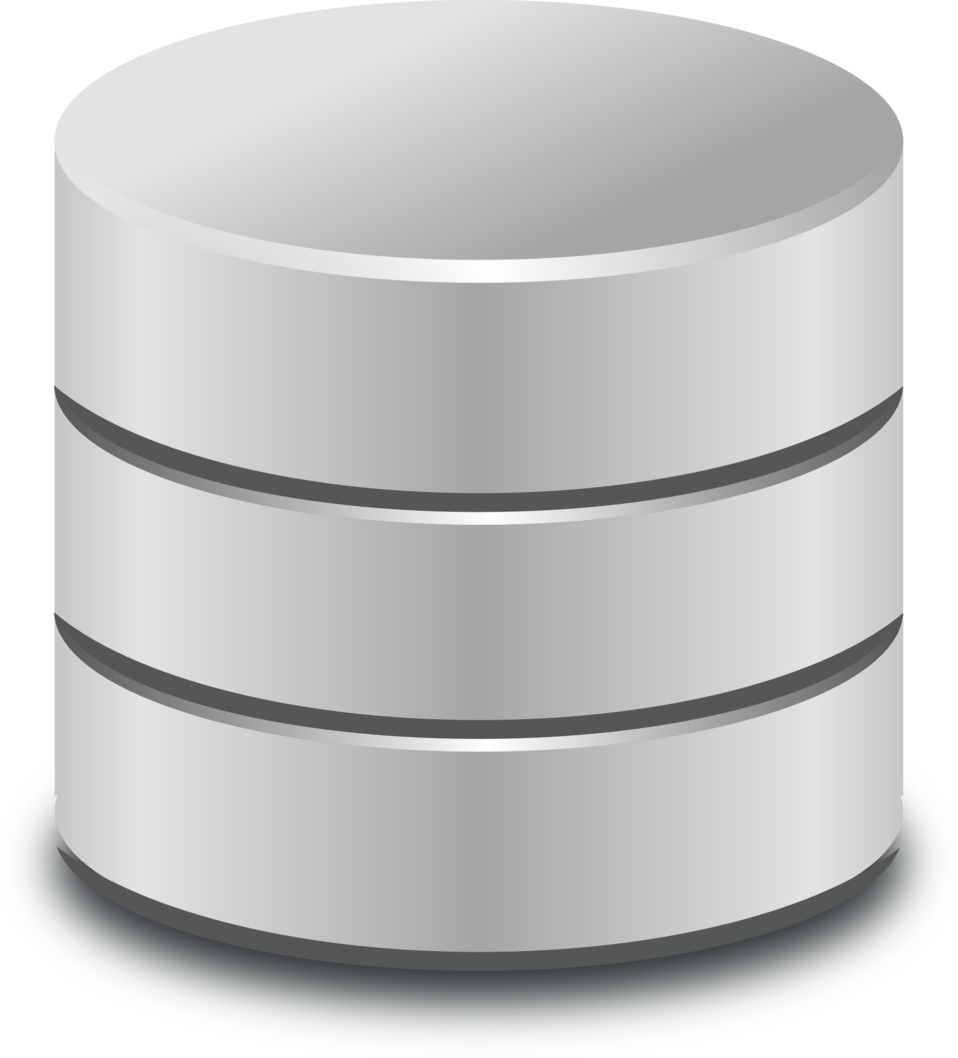}
        };
        \node[minimum size=1.5cm,label={[align=center]below:\ding{195} collect\\causes}] at (0, -3) (values) {
            \includegraphics[width=2cm,height=2cm]{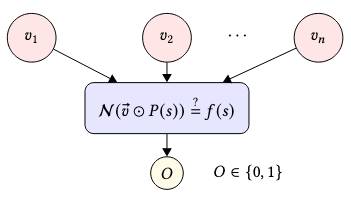}
        };
        \node[minimum size=2cm,label={[align=center]below:\ding{196} insert\\causes}] at (3, -3) (insert) {
            \includegraphics[scale=0.2]{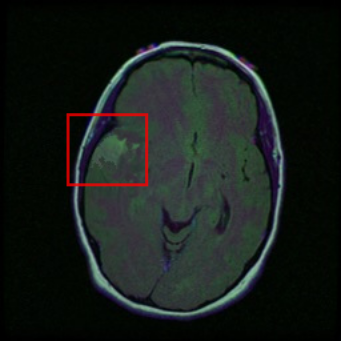}
        };
        \node[minimum size=2cm,label={[align=center]below:\ding{197} query\\model}] at (6, -3) (check) {
            \includegraphics[width=2cm,height=2cm]{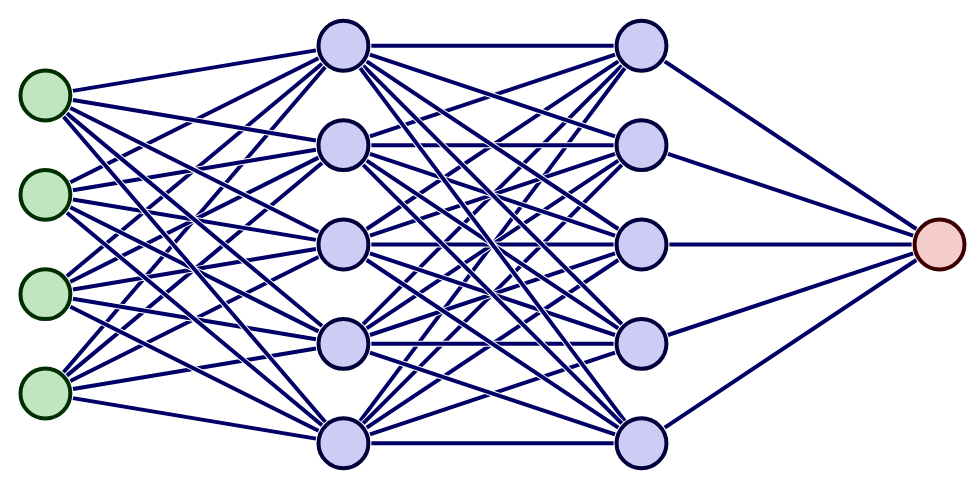}
        };
        \node[minimum size=2cm,align=center,label={[align=center]below:\ding{198} final\\counterfactual}] at (9, -3) (final) {
            \includegraphics[scale=0.2]{images/TCGA_DU_6408_19860521_21/se.jpg}
        };

        \draw[arrow] (start) -- (counterfactual);
        \draw[arrow] (counterfactual) -- (nearby);
        \draw[arrow] (nearby) -- (extract);
        \draw[arrow] (values) -- (insert);
        \draw[arrow] (check) -- (final);

        \draw[-{Triangle[reversed]},color=lightblue,line width=1.3mm] (extract) -- ++(0, -1.3) |- ++(-9, 0) |- ++(0, -0.3);

        \path[draw,-Triangle,line width=1.3mm,color=lightblue] (insert) to[bend right] (check);
        \path[draw,-Triangle,line width=1.3mm,color=lightblue] (check) to[bend right] (insert);
        
    \end{tikzpicture}
    \caption{Our workflow: \ding{192} an image classified as healthy; \ding{193} search for nearby images (mediated by some distance function); \ding{194} their causal explanations are pre-computed and saved in a database; \ding{195} extract causal values from the explanations; \ding{196}-\ding{197} main loop: insert causal values and query the model; finally \ding{198} an approximately minimal counterfactual generated by the classifier itself.}
    \label{fig:workflow}
\end{figure*}

Although both sufficient and complete explanations can be applied to positively and negatively classified images alike (\cite{kelly2025explaining}), they are less insightful for negative classifications (which indicate the \emph{absence} of a tumor). The reason is that these explanations rely on masking parts of the image to identify the location of a tumor, and thus they are less appropriate for an image where no tumor was identified to being with (see \Cref{sec:related}). Therefore negatively classified images require invoking a different type of well-known causal explanation, namely a counterfactual explanation. In the context of images, this means constructing counterfactual images.

Our approach to counterfactual image generation builds on the idea of viewing a classifier as a causal model, that allowed us to invoke well-defined notions of causal explanations in \Cref{sec:Background}. A counterfactual is a claim about what would have happened if some
occurrence were different, all else being equal. Concretely, it offers answers to queries of the form: ``Given that we observed $\vec{X} = \vec{x}$ that led to $O = o$, what would $O$ have been had some $\vec{X}' \subseteq \vec{X}$ been set to $\vec{x}'$''. Equivalently, using our previous notation, we are asking whether  $\mathcal{N}(\vec{x}'') = \mathcal{N}(\vec{x})$, where $\vec{x}''$ agrees with $\vec{x}'$ on $\vec{X}'$ and agrees with $\vec{x}$ elsewhere.

In principle, a {\em counterfactual} explanation of an image simply consists of any image that changes the classification, but the quality of a counterfactual explanation is determined by how {\em similar}, or close, the new image is to the original one~\cite{mertes2022ganterfactual}. Our method uses sufficient and complete explanations to systematically generate counterfactual explanations that are very close to the original image. \Cref{fig:workflow} shows a schematic of the workflow.

\subsection{Problem Statement}
Given an image that is originally classified as healthy, we ask the question ``What minimal causal evidence does the \emph{classifier itself} consider sufficient to change  its prediction to diseased?''. 
While \textbf{clinical plausibility} is important when counterfactuals are intended to model real-world disease processes, it is not a prerequisite for model auditing. Auditing seeks to understand the classifier's decision rather than the underlying pathology. By focusing on classifier-derived causal evidence rather than clinical realism, our approach provides a more direct probe of the model's decision boundary.

\begin{algorithm}[t]
  \caption{CausalCounterfactual $(\mathcal{N}, x, \mathcal{P}, \mathcal{P}_{\text{exp}}, d, M, t)$}
  \label{algo:gangone}
  \begin{flushleft}
    \textbf{INPUT:}\,\, a classifier $\mathcal{N}$, a negative image $x$, a set $\mathcal{P}$ of positive 
    images, a set $\mathcal{P}_{\text{exp}}$ of causal explanations, one for each member of $\mathcal{P}$, a distance function $d$, a set of masks $M$, a distance threshold value $t$;\\
    \textbf{OUTPUT:}\,\, a transformed image $x'$ or $\emptyset$
  \end{flushleft}
  \begin{algorithmic}[1]
    \STATE $p' \leftarrow$ arg min$_{p \in \mathcal{P}} \; d(x,p) $ 
    \STATE $E_p' \leftarrow$ get explanation of $p'$ from $\mathcal{P}_{\text{exp}}$
    \STATE $V_p' \leftarrow$ get\_actual\_values($E_p',p')$ 
    \STATE $C_p'$ $\leftarrow$ calculate\_centroid$(E_p')$
    \FOR{$m \in M$}
        \STATE $x' \gets$ apply $m$ to $x$ on location $C_p'$
        \WHILE{$d(x,x') < t$}
            \STATE $x' \gets $ update\_values\_under\_mask($x', V_p')$
            \IF{$\mathcal{N}(x')=1$}
                \STATE $x' \gets$ smooth($x'$)
                \RETURN $x'$
            \ENDIF 
        \ENDWHILE
    \ENDFOR
    \RETURN $\emptyset$
  \end{algorithmic}
\end{algorithm}

\subsection{Our Solution} 
Existing counterfactual methods explain a classifier using a second learned representation (a GAN latent space, diffusion latent space, etc.). Our method explains a classifier using evidence extracted from the classifier itself.
\Cref{algo:gangone} outlines our approach. 
The algorithm takes as input the reference set of positive images, $\mathcal{P}$, and its corresponding set of explanations, $\mathcal{P}_{\text{exp}}$. $\mathcal{P}$ can be obtained from the dataset used to train the classifier. $\mathcal{P}_{\text{exp}}$ is precomputed by \rex using the same model, $\mathcal{N}$, as in the main algorithm. 
We start by finding, in the given set of positive images $\mathcal{P}$, the image $p'$ that is closest to the target image $x$ (Line 1). This is mediated by a given distance function $d$. 

We find the matching explanation for this positive image (Line 2) in $\mathcal{P}_{exp}$. The explanation is given in the form of a binary mask, $E_p'$, defined over a matrix of the same size and shape as the image $p'$, and indicates which pixels are part of an explanation. We apply the mask over the positive image $p'$ to obtain the actual values of the pixels selected by $E'_p$. We take these actual values, remove duplicates, and sort them by their values, giving us an ordered array of values, $V'_p$. These values are what will be used to generate candidate perturbations. We need to choose a location in $x$ as a mutation target (Line 4). For our experiments, we use the centroid location of the abnormality in $p'$, but equally, a clinician could select a location and pass this on as an additional parameter to our algorithm.

The main loop of~\Cref{algo:gangone} starts on Line 5. We start by applying a user-defined binary \emph{mutation mask} $m$ over the chosen location in $x$, giving us $x'$. The mutant mask can be any initial size or shape (we use a simple rectangle in our experiments), and limits which pixels can be perturbed.
We move each value $v$ under $m$ one step closer to its closest value in $V'_p$. The effect of this is to bring pixels in $x'$ to pixel values known to be part of an explanation for $p'$. If this minimal shift does not change the outcome, we increase the step taken in $V'_p$, resulting in a new image with values further away from the original $x$, but with the same total number of altered pixels. The loop continues as long as $x'$ remains within the distance threshold $t$ from $x$. If we cannot change the classification of $x$ using the mask $m$ and values $V'_p$, we use a bigger mask $m'$. Line 10 is an optional smoothing calculation: the successful mask, $m$, might introduce hard lines or unnatural shapes, depending on initial choices; `smooth' attempts to reduce these artifacts. In our experiments, we start from rectangular masks and use a segmentation-based approach to smooth the outlines.

The algorithm always terminates, as both $V'_p$ and $M$ are finite. In the event that no combination of mask size and value shift changes the classification, we return the empty set. The overall computational complexity is $O(|M| \times |V|)$. Our algorithm includes a sort in Line 3, with the standard complexity $O(|V|\log{|V|})$, but this is dominated by $O(|M| \times |V|)$ in all but degenerate cases. 
The set of causal explanations, $\mathcal{P}_{\text{exp}}$, is precomputed once for a trained classifier using \rex during
pre-processing, hence there is no need to compute them for each counterfactual generation. The computational complexity of running \rex is given in \Cref{sec:rex}.

\section{Evaluation}\label{sec:eval}

This section evaluates our approach for generating counterfactual explanations in two case studies. The evaluation concentrates on the two main criteria for counterfactual images: 1) Correctness: the classifier should predict that the counterfactual image belongs to a different class than the original. 2) Similarity: the counterfactual should be as similar as possible to the original image. 
For the distance function $d$ in our algorithm implementation, we used the L2 norm. To compute the centroid of the explanation mask, we took the median location of the explanation. Our smoothing operator uses the SLIC segmentation algorithm~\cite{achanta2010slic}. The complete code for the evaluation and the implementation of our approach, including details on the data preprocessing, are available in~\cite{supplementarymaterial}.

\subsection{Case Studies}
The first case study was a dataset of brain MRI data obtained from The Cancer Imaging Archive~\cite{buda2019association}. This curated dataset of $110$ pre-operative patients with low grade gliomas (LGG) was gathered from $5$ US institutions. All the images are axial slices of the brain and include fluid-attenuated-inversion recovery (FLAIR) sequence, while $101$ also have pre-contrast sequence and $104$ have post-contrast sequence images. Each patient had between $20$ and $88$ slices taken, with a total of $3,929$ images. Each image contains $3$ channels, one for each of the three sequences. Following~\cite{legastelois2023challenges}, we used their pre-trained CNN classifier based on the ResNet50 architecture and did not perform any additional classifier training. Consistent with their experimental setup, the dataset was split at the patient level, with 80\% of patients used for training and 20\% for testing.

The second case study was a dataset of skin lesion images from The International Skin Imaging Collaboration (ISIC) archive~\cite{gutman2016skin}. The dataset includes a representative set of both malignant and benign skin lesion images. The dataset contains $1250$ images, partitioned into train and test before release. For the classifier, we took a trained attention-based CNN based on the VGG architecture from~\cite{yan2019melanoma}, and did not perform any additional classifier training.

In our evaluation, we considered all images that the classifier predicted as negative, including both true and false negatives. This choice reflected the intended clinical use, where a clinician seeks an explanation of the model, where the ground-truth is not yet established. This yielded $560$ tested images for the brain MRI dataset and $277$ for the ISIC skin lesion dataset.

\subsection{Evaluation Metrics}
A counterfactual image explains a classification only if it remains sufficiently similar, or close, to the original image. 
A sensible distance metric should take into account two dimensions: the set of pixels on which they disagree, and/or the disagreement between the pixel values. The first dimension gives us local explanations, allowing the identification of a specific, small, part of the input that is responsible for the change in classification. The second dimension gives us insight into how substantial the change was. The trade-off between both dimensions depends on the distance metric used and can be domain-dependent.  Therefore, we evaluated the quality of the counterfactual image under a distance constraint, using several distance metrics. More specifically, we used the distance-constrained success criterion, where for a given threshold $t$ original image $x$ and generate image $x'$, $$\success(x,x',t) = \mathbb{I}[\mathcal{N}(x) \ne \mathcal{N}(x') \land d(x,x') < t].$$ Simply put, under this success criterion, a counterfactual is successful if it changes the original prediction of the classifier, and its distance from the original image is lower than $t$.

%paper that discuss distance meetrics and counterfactuals - https://arxiv.org/pdf/2103.04244
We evaluated distances using the L1 and L2 norms, Structural Similarity Index Measure (SSIM)~\cite{1284395}, and the Learned Perceptual Image Patch Similarity (LPIPS) metric~\cite{zhang2018perceptual}. The latter better aligns with human perceptions of image similarity. In our implementation of the algorithm, we used the L2 norm as the distance function to find the closest positive image in the reference dataset $\mathcal{P}$. This choice was deliberate, since our method aims to be fully transparent, and incorporating LPIPS, which is black-box, would undermine this goal. However, we stress that our approach is general and can accommodate any distance metric, interpretable or not. We selected L2 over other interpretable metrics because it yielded slightly better performance. We still include all distance metrics as a part of the evaluation to align with prior related work~\cite{mertes2022ganterfactual,zigutyte_2024}.

In addition, to assess whether the different approaches to counterfactual generation map the population to a similar region of the input space, we computed all pairwise inter-image distances before and after transformation and compared the resulting distance distributions (\Cref{fig:distance}).

\subsection{Comparison}
We evaluated our approach using two variants: one using sufficient explanations (\textbf{\SE}), and one using complete explanations (\textbf{\CE}). To evaluate our contribution, we compared our approach with two state-of-the-art methods for generating counterfactual explanations for medical images: 1) GANterfactual, a generative adversarial learning (\textbf{\GAN}) approach~\cite{mertes2022ganterfactual}, and 2) MoPaDi, a diffusion model (\textbf{\DM}) approach~\cite{zigutyte_2024,Preechakul_2022_CVPR}. Both GANterfactual and MoPaDi were trained using the default parameters provided in their publicly available implementations. We also evaluated a naive approach that creates a counterfactual image by taking the explanation of the closest positive image in the training data and inserting it onto the original image. We show the results of this approach for both sufficient and complete explanations, which we denote naive sufficient explanations (\textbf{\NPS}) and naive complete explanations (\textbf{\NPC}), respectively.
All methods were given access to the same training data. The generative approaches (\textbf{\GAN} and \textbf{\DM}) used the full training set during model training, whereas the causal approaches (\textbf{\SE}, \textbf{\CE}, \textbf{\NPS}, and \textbf{\NPC}) used the positively classified training samples to construct the precomputed reference database of causal explanations used for counterfactual generation.

\subsection{Results}\label{sec:results}

\input{results_plots}

\input{brain_plot} 

\Cref{fig:success} presents the distance-constrained success rate of the methods for both datasets. To determine a meaningful upper bound on acceptable distances, we selected the cutoff based on the variability in its respective dataset. Specifically, we computed the distribution of pairwise distances between images in the dataset and selected the 99$^{th}$ percentile as the maximal distance threshold. Distances beyond this point correspond to image pairs that are already extremely dissimilar within the dataset itself, making it unlikely that a counterfactual exceeding this bound would remain usefully similar to the original image from which it was generated.

On the brain MRI dataset, the \textbf{\CE} approach outperformed the other approaches at lower distance thresholds, achieving an 85\% success rate with distance thresholds of 0.25 in LPIPS and 0.2 in SSIM. Additionally, the diffusion model (DM) method differed from the causal approaches, producing successful counterfactuals but which were significantly farther from their original images. The GAN approach generated correct counterfactuals for most of the dataset, but they differed substantially from the original images. In the case of the brain MRI use case with L1 and L2, none of the counterfactuals were at a distance less than the maximal threshold, and therefore, its distance-constrained success rate remains 0 in the examined range.

For the skin lesion dataset, the  \textbf{\CE} approach again outperformed other approaches at lower distance thresholds, achieving a 67\% success rate at distance thresholds of 0.48 in LPIPS and 0.21 in SSIM. At higher distance thresholds, \textbf{\NPC}, \textbf{\GAN}, and \textbf{\DM} achieve higher success rates. However, this improvement is driven by transformations that substantially alter the original image (examples of such transformations are shown in \Cref{fig:result_images_brain}). That is, it is always possible to perturb an image sufficiently to change the classification. However, larger changes reduce the practical usefulness of the explanation. This accounts for the poor performance of these methods at lower distance thresholds.

\Cref{fig:distance} shows the distribution of the pairwise inter-image distances for the original images (before transformation) and after transformation of each method. To ensure a fair comparison, we restrict the dataset to the intersection of images for which all methods successfully generated a valid counterfactual, so that all statistics are computed over an identical evaluation set. All causal approaches preserved the distribution of the original dataset in both use cases, whereas the \GAN approach generated counterfactuals with pairwise distances that were significantly lower than those of the original dataset. This indicates that the GAN model converged to a narrow transformation pattern, producing counterfactuals that are highly similar to each other but highly dissimilar from the original data distribution. This is evidence that the GAN model has learned a simple shortcut, transforming all images to very similar counterfactuals (visible as the sharp peaks in \Cref{fig:distance}). This may be exacerbated by shortcut learning in the original classifier, which is common in medical models which are often trained in data-sparse regimes. The \DM approach showed a similar trend in the skin lesion dataset. However, for the brain MRI dataset, the \DM generated counterfactuals with pairwise distances that were higher than those of the original dataset.

\subsection{Limitations}

% In this paper, we focus on transforming negatively classified images (`no-tumor') into positively classified ones (`tumor'). This choice is driven by the fact that pathology detection naturally provides identifiable positive evidence from which causal explanations or useful feature-attribution scores can be extracted. In contrast, negatively classified images correspond to the absence of pathology, for which meaningful causal explanations are often unavailable or significantly less informative~\cite{kelly2025explaining}. As a result, explanations derived from negative images are generally not useful for constructing perturbations that transform positive images into negative ones in this setting. At the same time, this limitation also highlights why counterfactual explanations are particularly valuable for negatively classified samples, where standard causal explanations and feature attribution scores are less effective.

% \tom{relavant to use case where local changes is}

An important caveat of our approach is its dependence on the diversity of the reference set and the corresponding explanations. The method will likely benefit from a reference dataset with a broad range of pathological appearances and imaging conditions. If certain patterns are underrepresented, the space of available perturbations may become limited. Nevertheless, this dependency is less restrictive than those imposed by generative methods. Generative models must learn a full image distribution and, therefore, typically require large, diverse datasets to produce viable results. In contrast, our approach reuses localized causal evidence directly from existing images and does not require training an additional model. As a result, the data requirements are substantially smaller.

\section{Related Work}\label{sec:related}
Counterfactual explanations have become an important approach for providing actionable explanations and satisfying emerging expectations for AI transparency~\cite{wachter2017counterfactual,verma2020counterfactual,guidotti2018survey}. In computer vision, most counterfactual methods generate an alternative image that would be assigned a different label by a classifier. A dominant line of work uses generative models to perform image-to-image translation. Early approaches employed CycleGANs to change class labels without requiring paired training examples~\cite{zhu2017unpaired}, a property that is particularly attractive in medical imaging where paired pathological and healthy images are rarely available. Building on this idea, GANterfactual~\cite{mertes2022ganterfactual} used GAN-based image translation to produce medical counterfactual explanations, while~\cite{singla2023explaining} introduced a conditional GAN framework designed to preserve rare but clinically important image details. 

More recent work has largely shifted towards diffusion-based generators.~\cite{zigutyte_2024} and~\cite{atad2024counterfactual} exploit diffusion models to generate medically plausible counterfactuals, while~\cite{liu2024controllable} and~\cite{oh2022learn} learn dedicated generative mechanisms for controllable counterfactual image synthesis in medical settings. Beyond image generation,~\cite{xia2024mitigating} employ a deep structural causal model to manipulate demographic and clinical factors in chest radiographs, enabling counterfactual analyses of attributes such as race and sex. Despite their methodological differences, these approaches share a common characteristic: they require training an additional generative or structural model whose learned representation mediates the explanation process. This reliance on a second model introduces a potential tension for explainability. The resulting counterfactual may reflect not only the classifier under inspection but also the inductive biases of the generator itself. Consequently, it can be difficult to disentangle whether a generated image reflects the classifier's actual decision boundary, the generative model's image prior, or optimization artifacts~\cite{cohen2021gifsplanation}. Furthermore, most existing methods prioritize visual realism and clinical plausibility. While desirable for data generation or simulation, realism is not necessarily synonymous with explanation: a highly realistic image may obscure the specific evidence that the classifier uses when making a prediction~\cite{bhusalface,singla2023explaining}.

The closest work to ours is~\cite{goyal19a}, who generate counterfactual visual explanations by copying image regions from a ``distractor'' image into a query image. Their method demonstrates that small localized edits can alter classifier predictions, but the transferred content is selected through an optimization procedure and is not tied to features that are known to be causally responsible for the classifier's decision. Consequently, it remains unclear why a particular perturbation succeeds. Our approach differs from both generative and retrieval-based counterfactual methods in a fundamental way. Rather than learning a latent representation of the data or searching for effective pixel substitutions, we construct counterfactuals directly from causal evidence extracted from the classifier itself. To our knowledge, this is the first image-counterfactual framework in which causal explanations are not merely used to interpret counterfactuals after generation, but serve as the primary representation from which counterfactuals are constructed.

\section{Conclusion}

We explore the nature of counterfactual images from the perspective of the classifier. By calculating the information change required to move `healthy' to `diseased' images using only the classifier itself, we are able to provide `what-if' images to clinicians tasked with understanding a model's modes of failure. Our method induces a change in classification using far smaller changes compared to state-of-the-art generative models, resulting in counterfactuals closer to the original image. Future work will assess whether clinicians do indeed find our counterfactual images useful for auditing models.

{
    \small
    \bibliographystyle{ieeenat_fullname}
    \bibliography{refs}
}

\end{document}

%% file: preamble.tex
\usepackage{pgfplots}
\usepgfplotslibrary{groupplots,dateplot}
\usetikzlibrary{patterns,shapes,arrows.meta,bending,backgrounds}
\pgfplotsset{compat=newest}

\usepackage{pgf}
\usepackage{pgfplots}
\usepackage{xspace}
\pgfplotsset{width=5cm,compat=1.9}
\pgfplotsset{every tick label/.append style={font=\tiny}}
\usepackage{graphicx}
\usepackage{booktabs}
\usepackage{multirow}
\usepackage{pifont}

\usepackage{xspace}
\usepackage{amsmath}
\usepackage{amssymb}
\usepackage{mathtools}
\usepackage{amsthm}
\usepackage{nicefrac}
\usepackage{amsfonts}
\usepackage{pifont}
\usepackage{floatflt}
\usepackage{url}
\usepackage{dsfont}

\usepackage{tikz}
\usetikzlibrary{calc, positioning, shadows, backgrounds}
\usepackage{multirow}
\usepackage{booktabs}
\usepackage{tabularx}
\newcolumntype{L}{>{\raggedright\arraybackslash}X}

\usepackage{algorithm}
\usepackage{algorithmic}

\theoremstyle{plain}
\theoremstyle{definition}
\newtheorem{definition}{Definition}
\theoremstyle{remark}
\theoremstyle{plain}

\makeatletter
\DeclareRobustCommand*\cal{\@fontswitch\relax\mathcal}
\makeatother

\newcommand{\xai}{\textsc{xai}\xspace}

\newcommand{\lime}{\textsc{lime}\xspace}
\newcommand{\plainshap}{{\textsc{shap}}\xspace}

\newcommand{\rex}{\textsc{\sc r}e{\sc x}\xspace}

\newcommand{\commentout}[1]{}

\newcommand{\dfn}{\begin{definition}}
\newcommand{\edfn}{\end{definition}}

\newcommand{\?}{\stackrel{?}{=}}

%% file: results_plots.tex
\begin{figure*}[t]
    \centering
    \begin{subfigure}[t]{0.48\textwidth}
        \centering
    \scalebox{0.8}{%
    \begin{tikzpicture}
   
        \begin{groupplot}[group style={rows=4,columns=2},
        view={0}{90},
        width=5cm,height=3.2cm]
        \nextgroupplot[xlabel={LPIPS (MRI)},ylabel={Success rate},ymin=0,ymax=1,xmin=0,xmax=0.8,ymajorgrids,xmajorgrids]
        \pgfplotstableread[col sep=comma]{plots_data/brain/merged_evaluation_results_lpips.csv}\datatable
        \addplot[line width=0.3mm, color=blue!70] table[x=threshold, y=se] {\datatable};
        \addplot[line width=0.3mm, color=green!70] table[x=threshold, y=ne] {\datatable};
        \addplot[line width=0.3mm, color=red!70] table[x=threshold, y=nps] {\datatable};
        \addplot[line width=0.3mm, color=orange!70] table[x=threshold, y=npn] {\datatable};
        \addplot[line width=0.3mm, color=black!80] table[x=threshold, y=gan] {\datatable};
        \addplot[line width=0.3mm, color=brown!100] table[x=threshold, y=diffusion] {\datatable};

        \nextgroupplot[xlabel={LPIPS (ISIC)},ymin=0,ymax=1,xmin=0,xmax=0.8,ymajorgrids,xmajorgrids]
        \pgfplotstableread[col sep=comma]{plots_data/isic/merged_evaluation_results_lpips.csv}\datatable
        \addplot[line width=0.3mm, color=blue!70] table[x=threshold, y=se] {\datatable};
        \addplot[line width=0.3mm, color=green!70] table[x=threshold, y=ne] {\datatable};
        \addplot[line width=0.3mm, color=red!70] table[x=threshold, y=nps] {\datatable};
        \addplot[line width=0.3mm, color=orange!70] table[x=threshold, y=npn] {\datatable};
        \addplot[line width=0.3mm, color=black!80] table[x=threshold, y=gan] {\datatable};
        \addplot[line width=0.3mm, color=brown!100] table[x=threshold, y=diffusion] {\datatable};

        \nextgroupplot[xlabel={1-SSIM (MRI)},ylabel={Success rate},ymin=0,ymax=1,xmin=0,xmax=0.81,ymajorgrids,xmajorgrids]
        \pgfplotstableread[col sep=comma]{plots_data/brain/merged_evaluation_results_ssim.csv}\datatable
        \addplot[line width=0.3mm, color=blue!70] table[x=threshold, y=se] {\datatable};
        \addplot[line width=0.3mm, color=green!70] table[x=threshold, y=ne] {\datatable};
        \addplot[line width=0.3mm, color=red!70] table[x=threshold, y=nps] {\datatable};
        \addplot[line width=0.3mm, color=orange!70] table[x=threshold, y=npn] {\datatable};
        \addplot[line width=0.3mm, color=black!80] table[x=threshold, y=gan] {\datatable};
        \addplot[line width=0.3mm, color=brown!100] table[x=threshold, y=diffusion] {\datatable};

        \nextgroupplot[xlabel={1-SSIM (ISIC)},ymin=0,ymax=1,xmin=0,xmax=0.81,ymajorgrids,xmajorgrids]
        \pgfplotstableread[col sep=comma]{plots_data/isic/merged_evaluation_results_ssim.csv}\datatable
        \addplot[line width=0.3mm, color=blue!70] table[x=threshold, y=se] {\datatable};
        \addplot[line width=0.3mm, color=green!70] table[x=threshold, y=ne] {\datatable};
        \addplot[line width=0.3mm, color=red!70] table[x=threshold, y=nps] {\datatable};
        \addplot[line width=0.3mm, color=orange!70] table[x=threshold, y=npn] {\datatable};
        \addplot[line width=0.3mm, color=black!80] table[x=threshold, y=gan] {\datatable};
        \addplot[line width=0.3mm, color=brown!100] table[x=threshold, y=diffusion] {\datatable};

        \nextgroupplot[xlabel={L1 (MRI)},ylabel={Success rate},ymin=-0.1,ymax=1,xmin=0,xmax=43.2,ymajorgrids,xmajorgrids]
        \pgfplotstableread[col sep=comma]{plots_data/brain/merged_evaluation_results_l1.csv}\datatable
        \addplot[line width=0.3mm, color=blue!70] table[x=threshold, y=se] {\datatable};
        \addplot[line width=0.3mm, color=green!70] table[x=threshold, y=ne] {\datatable};
        \addplot[line width=0.3mm, color=red!70] table[x=threshold, y=nps] {\datatable};
        \addplot[line width=0.3mm, color=orange!70] table[x=threshold, y=npn] {\datatable};
        \addplot[line width=0.3mm, color=black!80] table[x=threshold, y=gan] {\datatable};
        \addplot[line width=0.3mm, color=brown!100] table[x=threshold, y=diffusion] {\datatable};

        \nextgroupplot[xlabel={L1 (ISIC)},ymin=-0.1,ymax=1,xmin=0,xmax=136.3,ymajorgrids,xmajorgrids]
        \pgfplotstableread[col sep=comma]{plots_data/isic/merged_evaluation_results_l1.csv}\datatable
        \addplot[line width=0.3mm, color=blue!70] table[x=threshold, y=se] {\datatable};
        \addplot[line width=0.3mm, color=green!70] table[x=threshold, y=ne] {\datatable};
        \addplot[line width=0.3mm, color=red!70] table[x=threshold, y=nps] {\datatable};
        \addplot[line width=0.3mm, color=orange!70] table[x=threshold, y=npn] {\datatable};
        \addplot[line width=0.3mm, color=black!80] table[x=threshold, y=gan] {\datatable};
        \addplot[line width=0.3mm, color=brown!100] table[x=threshold, y=diffusion] {\datatable};

        \nextgroupplot[xlabel={L2 (MRI)},ylabel={Success rate},ymin=-0.1,ymax=1,xmin=0,xmax=4000,ymajorgrids,xmajorgrids]
        \pgfplotstableread[col sep=comma]{plots_data/brain/merged_evaluation_results_l2.csv}\datatable
        \addplot[line width=0.3mm, color=blue!70] table[x=threshold, y=se] {\datatable};
        \addplot[line width=0.3mm, color=green!70] table[x=threshold, y=ne] {\datatable};
        \addplot[line width=0.3mm, color=red!70] table[x=threshold, y=nps] {\datatable};
        \addplot[line width=0.3mm, color=orange!70] table[x=threshold, y=npn] {\datatable};
        \addplot[line width=0.3mm, color=black!80] table[x=threshold, y=gan] {\datatable};
        \addplot[line width=0.3mm, color=brown!100] table[x=threshold, y=diffusion] {\datatable};

        \nextgroupplot[xlabel={L2 (ISIC)},ymin=-0.1,ymax=1,xmin=0,xmax=2000,ymajorgrids,xmajorgrids,legend style={at={(-0.1, -1.0)},
        anchor=south},
        legend columns=-1]
        \pgfplotstableread[col sep=comma]{plots_data/isic/merged_evaluation_results_l2.csv}\datatable
        \addplot[line width=0.3mm, color=blue!70] table[x=threshold, y=se] {\datatable};
        \addlegendentry{\SE}
        \addplot[line width=0.3mm, color=green!70] table[x=threshold, y=ne] {\datatable};
        \addlegendentry{\CE}
        \addplot[line width=0.3mm, color=red!70] table[x=threshold, y=nps] {\datatable};
        \addlegendentry{\NPS}
        \addplot[line width=0.3mm, color=orange!70] table[x=threshold, y=npn] {\datatable};
        \addlegendentry{\NPC}
        \addplot[line width=0.3mm, color=black!80] table[x=threshold, y=gan] {\datatable};
        \addlegendentry{\GAN}
        \addplot[line width=0.3mm, color=brown!100] table[x=threshold, y=diffusion] {\datatable};
        \addlegendentry{\DM} 
        \end{groupplot}

    \end{tikzpicture}}
    \caption{The distance-constrained success rate of the examined methods for the brain MRI (left column) and ISIC skin lesion datasets (right column). The top row presents the results for the LPIPS metric for the two case studies, followed by the complement of SSIM (1-SSIM), L1, and L2.}%
    \label{fig:success}
    \end{subfigure}\hfill
    \begin{subfigure}[t]{0.48\textwidth}
    \centering
    \scalebox{0.8}{%
    \begin{tikzpicture}
        \begin{groupplot}[group style={rows=4,columns=2},
        view={0}{90},
        width=5cm,height=3.2cm]
        \nextgroupplot[xlabel={LPIPS (MRI)},ylabel={Density},ymin=0,ymax=6,ymajorticks=false,ymajorgrids,xmajorgrids,xmax=1]
        \addplot[line width=0.25mm,color=purple!70]
        table[col sep=comma, x=x, y=y]{plots_data/brain/merged_distances_histogram_original_lpips.csv};
        \addplot[line width=0.25mm,color=blue!70]
        table[col sep=comma, x=x, y=y]{plots_data/brain/merged_distances_histogram_se_lpips.csv};
        \addplot[line width=0.25mm,color=green!70]
        table[col sep=comma, x=x, y=y]{plots_data/brain/merged_distances_histogram_ne_lpips.csv};
        \addplot[line width=0.25mm,color=red!70]
        table[col sep=comma, x=x, y=y]{plots_data/brain/merged_distances_histogram_nps_lpips.csv};
        \addplot[line width=0.25mm,color=orange!70]
        table[col sep=comma, x=x, y=y]{plots_data/brain/merged_distances_histogram_npn_lpips.csv};
        \addplot[line width=0.25mm,color=black!80]
        table[col sep=comma, x=x, y=y]{plots_data/brain/merged_distances_histogram_gan_lpips.csv};
        \addplot[line width=0.25mm,color=brown!100]
        table[col sep=comma, x=x, y=y]{plots_data/brain/merged_distances_histogram_diffusion_lpips.csv};

        \nextgroupplot[xlabel={LPIPS (ISIC)},ymin=0,ymax=6,ymajorticks=false,ymajorgrids,xmajorgrids
        ]
        \addplot[line width=0.25mm,color=purple!70]
        table[col sep=comma, x=x, y=y]{plots_data/isic/merged_distances_histogram_original_lpips.csv};
        \addplot[line width=0.25mm,color=blue!70]
        table[col sep=comma, x=x, y=y]{plots_data/isic/merged_distances_histogram_se_lpips.csv};
        \addplot[line width=0.25mm,color=green!70]
        table[col sep=comma, x=x, y=y]{plots_data/isic/merged_distances_histogram_ne_lpips.csv};
        \addplot[line width=0.25mm,color=red!70]
        table[col sep=comma, x=x, y=y]{plots_data/isic/merged_distances_histogram_nps_lpips.csv};
        \addplot[line width=0.25mm,color=orange!70]
        table[col sep=comma, x=x, y=y]{plots_data/isic/merged_distances_histogram_npn_lpips.csv};
        \addplot[line width=0.25mm,color=black!80]
        table[col sep=comma, x=x, y=y]{plots_data/isic/merged_distances_histogram_gan_lpips.csv};
        \addplot[line width=0.25mm,color=brown!100]
        table[col sep=comma, x=x, y=y]{plots_data/isic/merged_distances_histogram_diffusion_lpips.csv};

        \nextgroupplot[xlabel={1-SSIM (MRI)},ylabel={Density},ymin=0,ymax=12,ymajorticks=false,ymajorgrids,xmajorgrids,xmax=1]
        \addplot[line width=0.25mm,color=purple!70]
        table[col sep=comma, x=x, y=y]{plots_data/brain/merged_distances_histogram_original_ssim.csv};
        \addplot[line width=0.25mm,color=blue!70]
        table[col sep=comma, x=x, y=y]{plots_data/brain/merged_distances_histogram_se_ssim.csv};
        \addplot[line width=0.25mm,color=green!70]
        table[col sep=comma, x=x, y=y]{plots_data/brain/merged_distances_histogram_ne_ssim.csv};
        \addplot[line width=0.25mm,color=red!70]
        table[col sep=comma, x=x, y=y]{plots_data/brain/merged_distances_histogram_nps_ssim.csv};
        \addplot[line width=0.25mm,color=orange!70]
        table[col sep=comma, x=x, y=y]{plots_data/brain/merged_distances_histogram_npn_ssim.csv};
        \addplot[line width=0.25mm,color=black!80]
        table[col sep=comma, x=x, y=y]{plots_data/brain/merged_distances_histogram_gan_ssim.csv};
        \addplot[line width=0.25mm,color=brown!100]
        table[col sep=comma, x=x, y=y]{plots_data/brain/merged_distances_histogram_diffusion_ssim.csv};

        \nextgroupplot[xlabel={1-SSIM (ISIC)},ymin=0,ymax=12,ymajorticks=false,ymajorgrids,xmajorgrids
        ]
        \addplot[line width=0.25mm,color=purple!70]
        table[col sep=comma, x=x, y=y]{plots_data/isic/merged_distances_histogram_original_ssim.csv};
        \addplot[line width=0.25mm,color=blue!70]
        table[col sep=comma, x=x, y=y]{plots_data/isic/merged_distances_histogram_se_ssim.csv};
        \addplot[line width=0.25mm,color=green!70]
        table[col sep=comma, x=x, y=y]{plots_data/isic/merged_distances_histogram_ne_ssim.csv};
        \addplot[line width=0.25mm,color=red!70]
        table[col sep=comma, x=x, y=y]{plots_data/isic/merged_distances_histogram_nps_ssim.csv};
        \addplot[line width=0.25mm,color=orange!70]
        table[col sep=comma, x=x, y=y]{plots_data/isic/merged_distances_histogram_npn_ssim.csv};
        \addplot[line width=0.25mm,color=black!80]
        table[col sep=comma, x=x, y=y]{plots_data/isic/merged_distances_histogram_gan_ssim.csv};
        \addplot[line width=0.25mm,color=brown!100]
        table[col sep=comma, x=x, y=y]{plots_data/isic/merged_distances_histogram_diffusion_ssim.csv};

        \nextgroupplot[xlabel={L1 (MRI)},ylabel={Density},ymin=0,ymax=0.28,ymajorticks=false,ymajorgrids,xmajorgrids,xmax=50]
        \addplot[line width=0.25mm,color=purple!70]
        table[col sep=comma, x=x, y=y]{plots_data/brain/merged_distances_histogram_original_l1.csv};
        \addplot[line width=0.25mm,color=blue!70]
        table[col sep=comma, x=x, y=y]{plots_data/brain/merged_distances_histogram_se_l1.csv};
        \addplot[line width=0.25mm,color=green!70]
        table[col sep=comma, x=x, y=y]{plots_data/brain/merged_distances_histogram_ne_l1.csv};
        \addplot[line width=0.25mm,color=red!70]
        table[col sep=comma, x=x, y=y]{plots_data/brain/merged_distances_histogram_nps_l1.csv};
        \addplot[line width=0.25mm,color=orange!70]
        table[col sep=comma, x=x, y=y]{plots_data/brain/merged_distances_histogram_npn_l1.csv};
        \addplot[line width=0.25mm,color=black!80]
        table[col sep=comma, x=x, y=y]{plots_data/brain/merged_distances_histogram_gan_l1.csv};
        \addplot[line width=0.25mm,color=brown!100]
        table[col sep=comma, x=x, y=y]{plots_data/brain/merged_distances_histogram_diffusion_l1.csv};

        \nextgroupplot[xlabel={L1 (ISIC)},ymin=0,ymax=0.16,ymajorticks=false,ymajorgrids,xmajorgrids
        ]
        \addplot[line width=0.25mm,color=purple!70]
        table[col sep=comma, x=x, y=y]{plots_data/isic/merged_distances_histogram_original_l1.csv};
        \addplot[line width=0.25mm,color=blue!70]
        table[col sep=comma, x=x, y=y]{plots_data/isic/merged_distances_histogram_se_l1.csv};
        \addplot[line width=0.25mm,color=green!70]
        table[col sep=comma, x=x, y=y]{plots_data/isic/merged_distances_histogram_ne_l1.csv};
        \addplot[line width=0.25mm,color=red!70]
        table[col sep=comma, x=x, y=y]{plots_data/isic/merged_distances_histogram_nps_l1.csv};
        \addplot[line width=0.25mm,color=orange!70]
        table[col sep=comma, x=x, y=y]{plots_data/isic/merged_distances_histogram_npn_l1.csv};
        \addplot[line width=0.25mm,color=black!80]
        table[col sep=comma, x=x, y=y]{plots_data/isic/merged_distances_histogram_gan_l1.csv};
        \addplot[line width=0.25mm,color=brown!100]
        table[col sep=comma, x=x, y=y]{plots_data/isic/merged_distances_histogram_diffusion_l1.csv};

        \nextgroupplot[xlabel={L2 (MRI)},ylabel={Density},ymin=0,ymax=0.01,ymajorticks=false,ymajorgrids,xmajorgrids,xmax=4000]
        \addplot[line width=0.25mm,color=purple!70]
        table[col sep=comma, x=x, y=y]{plots_data/brain/merged_distances_histogram_original_l2.csv};
        \addplot[line width=0.25mm,color=blue!70]
        table[col sep=comma, x=x, y=y]{plots_data/brain/merged_distances_histogram_se_l2.csv};
        \addplot[line width=0.25mm,color=green!70]
        table[col sep=comma, x=x, y=y]{plots_data/brain/merged_distances_histogram_ne_l2.csv};
        \addplot[line width=0.25mm,color=red!70]
        table[col sep=comma, x=x, y=y]{plots_data/brain/merged_distances_histogram_nps_l2.csv};
        \addplot[line width=0.25mm,color=orange!70]
        table[col sep=comma, x=x, y=y]{plots_data/brain/merged_distances_histogram_npn_l2.csv};
        \addplot[line width=0.25mm,color=black!80]
        table[col sep=comma, x=x, y=y]{plots_data/brain/merged_distances_histogram_gan_l2.csv};
        \addplot[line width=0.25mm,color=brown!100]
        table[col sep=comma, x=x, y=y]{plots_data/brain/merged_distances_histogram_diffusion_l2.csv};

        \nextgroupplot[xlabel={L2 (ISIC)},ymin=0,ymax=0.007,xmax=10000,ymajorticks=false,ymajorgrids,xmajorgrids,
        legend style={at={(-0.1, -1.0)},
        anchor=south},
        legend columns=-1,
        ]
        \addplot[line width=0.25mm,color=purple!70]
        table[col sep=comma, x=x, y=y]{plots_data/isic/merged_distances_histogram_original_l2.csv};
        \addlegendentry{Original}
        \addplot[line width=0.25mm,color=blue!70]
        table[col sep=comma, x=x, y=y]{plots_data/isic/merged_distances_histogram_se_l2.csv};
        \addlegendentry{\SE}
        \addplot[line width=0.25mm,color=green!70]
        table[col sep=comma, x=x, y=y]{plots_data/isic/merged_distances_histogram_ne_l2.csv};
        \addlegendentry{\CE}
        \addplot[line width=0.25mm,color=red!70]
        table[col sep=comma, x=x, y=y]{plots_data/isic/merged_distances_histogram_nps_l2.csv};
        \addlegendentry{\NPS}
        \addplot[line width=0.25mm,color=orange!70]
        table[col sep=comma, x=x, y=y]{plots_data/isic/merged_distances_histogram_npn_l2.csv};
        \addlegendentry{\NPC}
        \addplot[line width=0.25mm,color=black!80]
        table[col sep=comma, x=x, y=y]{plots_data/isic/merged_distances_histogram_gan_l2.csv};
        \addlegendentry{\GAN} 
        \addplot[line width=0.25mm,color=brown!100]
        table[col sep=comma, x=x, y=y]{plots_data/isic/merged_distances_histogram_diffusion_l2.csv};
        \addlegendentry{\DM} 
        
        \end{groupplot}
        
    \end{tikzpicture}}
    \caption{Distribution of the pairwise inter-image distances for the original images (before transformation) and after transformation of each method for the brain MRI (left column) and ISIC skin lesion datasets (right column). The top row presents the results for the LPIPS metric for the two case studies, followed by the complement of SSIM (1-SSIM), L1, and L2.}%
    \label{fig:distance}
    \end{subfigure}
\end{figure*}

%% file: brain_plot.tex
\begin{figure*}[t]
\centering
\begin{subfigure}{0.2\linewidth}\centering
\caption{Healthy}
\includegraphics[scale=0.27]{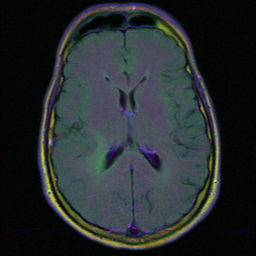}
\end{subfigure}
\begin{subfigure}{0.2\linewidth}\centering
\caption{Our Approach}
\includegraphics[scale=0.27]{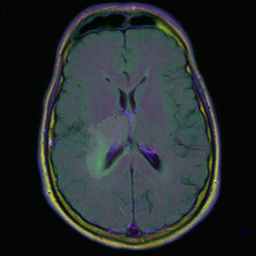}
\end{subfigure}
\begin{subfigure}{0.2\linewidth}\centering
\caption{Naive Patching}
\includegraphics[scale=0.27]{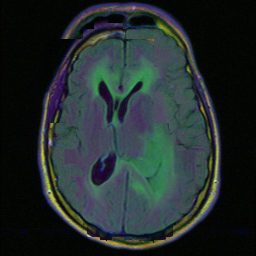}
\end{subfigure}
\begin{subfigure}{0.2\linewidth}\centering
\caption{Diffusion}
\includegraphics[scale=0.27]{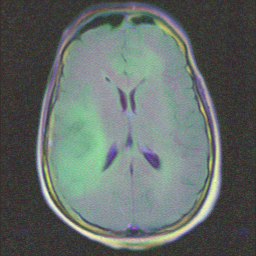}
\end{subfigure}\\
\vspace{5mm}
% row 2
\begin{subfigure}{0.2\linewidth}\centering
\includegraphics[scale=0.27]{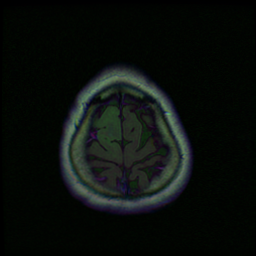}
\end{subfigure}
\begin{subfigure}{0.2\linewidth}\centering
\includegraphics[scale=0.27]{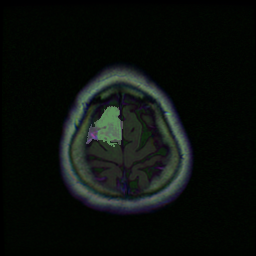}
\end{subfigure}
\begin{subfigure}{0.2\linewidth}\centering
\includegraphics[scale=0.27]{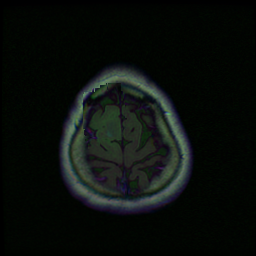}
\end{subfigure}
\begin{subfigure}{0.2\linewidth}\centering
\includegraphics[scale=0.27]{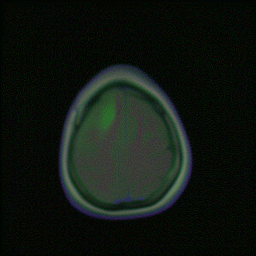}
\end{subfigure}\\
\vspace{5mm}
% row 3
\begin{subfigure}{0.2\linewidth}\centering
\includegraphics[scale=0.31]{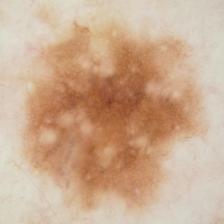}
\end{subfigure}
\begin{subfigure}{0.2\linewidth}\centering
\includegraphics[scale=0.31]{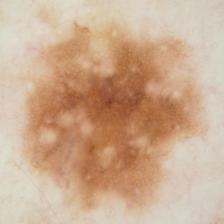}
\end{subfigure}
\begin{subfigure}{0.2\linewidth}\centering
\includegraphics[scale=0.31]{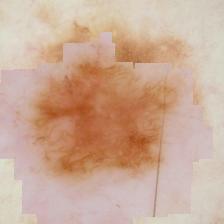}
\end{subfigure}
\begin{subfigure}{0.2\linewidth}\centering
\includegraphics[scale=0.31]{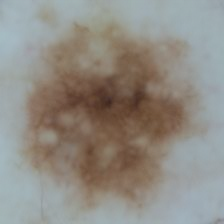}
\end{subfigure}\\
\vspace{5mm}
% row 4
\begin{subfigure}{0.2\linewidth}\centering
\includegraphics[scale=0.31]{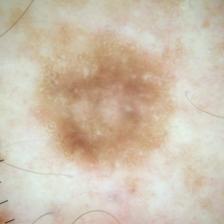}
\end{subfigure}
\begin{subfigure}{0.2\linewidth}\centering
\includegraphics[scale=0.31]{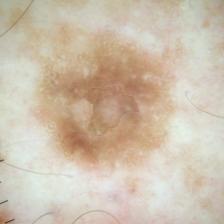}
\end{subfigure}
\begin{subfigure}{0.2\linewidth}\centering
\includegraphics[scale=0.31]{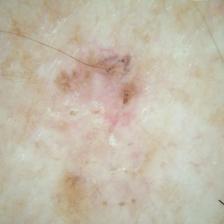}
\end{subfigure}
\begin{subfigure}{0.2\linewidth}\centering
\includegraphics[scale=0.31]{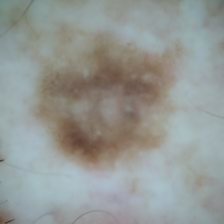}
\end{subfigure}

\caption{Our approach compared with other techniques on the two datasets. Column 1 shows samples classified as healthy. Using our method, we generate the images shown in column 2. If we apply `naive patching' we get the images in column 3. These are all classified as positive, but a large part of the original image has been replaced. In the skin lesion samples, the entire lesion is replaced. Finally, column 4 shows the output from the \DM. Although all these images are classified as positive, some images show a completely different sample, and in others, the changes are not local.
}
\label{fig:result_images_brain}
\end{figure*}

%% file: refs.bib
@ARTICLE{1284395,
  author={Zhou Wang and Bovik, A.C. and Sheikh, H.R. and Simoncelli, E.P.},
  journal={IEEE Transactions on Image Processing}, 
  title={Image quality assessment: from error visibility to structural similarity}, 
  year={2004},
  volume={13},
  number={4},
  pages={600-612},
  doi={10.1109/TIP.2003.819861}}

@article {zigutyte_2024,
	author = {{\v Z}igutyte, Laura and Lenz, Tim and Han, Tianyu and Hewitt, Katherine J. and Reitsam, Nic G. and Foersch, Sebastian and Carrero, Zunamys I. and Unger, Michaela and Pearson, Alexander T. and Truhn, Daniel and Kather, Jakob Nikolas},
	title = {Counterfactual Diffusion Models for Mechanistic Explainability of Artificial Intelligence Models in Pathology},
	elocation-id = {2024.10.29.620913},
	year = {2024},
	doi = {10.1101/2024.10.29.620913},
	publisher = {Cold Spring Harbor Laboratory},
	URL = {https://www.biorxiv.org/content/early/2024/11/03/2024.10.29.620913},
	eprint = {https://www.biorxiv.org/content/early/2024/11/03/2024.10.29.620913.full.pdf},
	journal = {bioRxiv}
}

@InProceedings{Preechakul_2022_CVPR,
    author    = {Preechakul, Konpat and Chatthee, Nattanat and Wizadwongsa, Suttisak and Suwajanakorn, Supasorn},
    title     = {Diffusion Autoencoders: Toward a Meaningful and Decodable Representation},
    booktitle = {Proceedings of the IEEE/CVF Conference on Computer Vision and Pattern Recognition (CVPR)},
    month     = {June},
    year      = {2022},
    pages     = {10619-10629}
}

@book{Pearl09a,
  author = {Pearl, Judea},
  edition = {2nd},
  publisher = {Cambridge University Press},
  title = {Causality: Models, Reasoning and Inference},
  year = 2009
}

@inproceedings{beckers22,
    title = {Causal Explanations and {XAI}},
    author = {Beckers, Sander},
    booktitle = {Proceedings of the First Conference on Causal Learning and
                 Reasoning},
    pages = {90--109},
    year = {2022},
    volume = {177},
    series = {Proceedings of Machine Learning Research},
    month = {11--13 Apr},
    publisher = {PMLR},
}

@misc{supplementarymaterial,
    author = {Anonymous Authors},
    title = {Counterfactual Explanations for Medical Imaging Classification
             using Causality --- Supplementary Material},
    howpublished = {\url{https://figshare.com/s/f82b2a8afe4b366ee07b}},
    year = {2026},
}

@inproceedings{zhang2018perceptual,
    title = {The Unreasonable Effectiveness of Deep Features as a Perceptual
             Metric},
    author = {Zhang, Richard and Isola, Phillip and Efros, Alexei A and
              Shechtman, Eli and Wang, Oliver},
    booktitle = {CVPR},
    year = {2018},
}

@techreport{achanta2010slic,
    title = {Slic superpixels},
    author = {Achanta, Radhakrishna and Shaji, Appu and Smith, Kevin and Lucchi,
              Aurelien and Fua, Pascal and S{\"u}sstrunk, Sabine and others},
    year = {2010},
    institution = {Technical report EPFL},
}

@inproceedings{yan2019melanoma,
    title = {Melanoma Recognition via Visual Attention},
    author = {Yan, Yiqi and Kawahara, Jeremy and Hamarneh, Ghassan},
    booktitle = {International Conference on Information Processing in Medical
                 Imaging},
    pages = {793--804},
    year = {2019},
    organization = {Springer},
}

@article{gutman2016skin,
    title = {Skin lesion analysis toward melanoma detection: A challenge at the
             international symposium on biomedical imaging (ISBI) 2016, hosted by
             the international skin imaging collaboration (ISIC)},
    author = {Gutman, David and Codella, Noel CF and Celebi, Emre and Helba,
              Brian and Marchetti, Michael and Mishra, Nabin and Halpern, Allan},
    journal = {arXiv preprint arXiv:1605.01397},
    year = {2016},
}

@article{mertes2022ganterfactual,
    title = {Ganterfactual—counterfactual explanations for medical non-experts
             using generative adversarial learning},
    author = {Mertes, Silvan and Huber, Tobias and Weitz, Katharina and Heimerl,
              Alexander and Andr{\'e}, Elisabeth},
    journal = {Frontiers in artificial intelligence},
    volume = {5},
    pages = {825565},
    year = {2022},
    publisher = {Frontiers Media SA},
}

@inproceedings{kelly2025explaining,
    title = {Explaining negative classifications of AI models in tumor diagnosis
             },
    author = {Kelly, David A and Chockler, Hana and Blake, Nathan},
    booktitle = {The 41st Conference on Uncertainty in Artificial Intelligence},
    year = {2025},
}

@inproceedings{legastelois2023challenges,
    title = {Challenges in Explaining Brain Tumor Detection},
    author = {Legastelois, Benedicte and Rafferty, Amy and Brennan, Paul and
              Chockler, Hana and Rajan, Ajitha and Belle, Vaishak},
    booktitle = {Proceedings of the First International Symposium on Trustworthy
                 Autonomous Systems},
    pages = {1--8},
    year = {2023},
}

@article{buda2019association,
    title = {Association of genomic subtypes of lower-grade gliomas with shape
             features automatically extracted by a deep learning algorithm},
    author = {Buda, Mateusz and Saha, Ashirbani and Mazurowski, Maciej A},
    journal = {Computers in biology and medicine},
    volume = {109},
    pages = {218--225},
    year = {2019},
    publisher = {Elsevier},
}

@inproceedings{liu2024controllable,
    title = {Controllable counterfactual generation for interpretable medical
             image classification},
    author = {Liu, Shiyu and Wang, Fan and Ren, Zehua and Lian, Chunfeng and Ma,
              Jianhua},
    booktitle = {International Conference on Medical Image Computing and
                 Computer-Assisted Intervention},
    pages = {143--152},
    year = {2024},
    organization = {Springer},
}

@article{singla2023explaining,
    title = {Explaining the black-box smoothly—a counterfactual approach},
    author = {Singla, Sumedha and Eslami, Motahhare and Pollack, Brian and
              Wallace, Stephen and Batmanghelich, Kayhan},
    journal = {Medical Image Analysis},
    volume = {84},
    pages = {102721},
    year = {2023},
    publisher = {Elsevier},
}

@inproceedings{xia2024mitigating,
    title = {Mitigating attribute amplification in counterfactual image
             generation},
    author = {Xia, Tian and Roschewitz, M{\'e}lanie and De Sousa Ribeiro, Fabio
              and Jones, Charles and Glocker, Ben},
    booktitle = {International Conference on Medical Image Computing and
                 Computer-Assisted Intervention},
    pages = {546--556},
    year = {2024},
    organization = {Springer},
}

@inproceedings{cohen2021gifsplanation,
    title = {Gifsplanation via latent shift: a simple autoencoder approach to
             counterfactual generation for chest x-rays},
    author = {Cohen, Joseph Paul and Brooks, Rupert and En, Sovann and Zucker,
              Evan and Pareek, Anuj and Lungren, Matthew P and Chaudhari, Akshay},
    booktitle = {Medical Imaging with Deep Learning},
    pages = {74--104},
    year = {2021},
    organization = {PMLR},
}

@article{wachter2017counterfactual,
    title = {Counterfactual explanations without opening the black box:
             Automated decisions and the GDPR},
    author = {Wachter, Sandra and Mittelstadt, Brent and Russell, Chris},
    journal = {Harv. JL \& Tech.},
    volume = {31},
    pages = {841},
    year = {2017},
    publisher = {HeinOnline},
}

@inproceedings{zhu2017unpaired,
    title = {Unpaired image-to-image translation using cycle-consistent
             adversarial networks},
    author = {Zhu, Jun-Yan and Park, Taesung and Isola, Phillip and Efros,
              Alexei A},
    booktitle = {Proceedings of the IEEE international conference on computer
                 vision},
    pages = {2223--2232},
    year = {2017},
}

@article{atad2024counterfactual,
title={Counterfactual Explanations for Medical Image Classification and Regression using Diffusion Autoencoder},
author={Atad, Matan and Schinz, David and Moeller, Hendrik and Graf, Robert and Wiestler, Benedikt and Rueckert, Daniel and Navab, Nassir and Kirschke, Jan S. and Keicher, Matthias},
journal={Machine Learning for Biomedical Imaging},
year={2024}
}

@book{Hal19,
    author = {Joseph Y. Halpern},
    title = {Actual Causality},
    year = 2016,
    publisher = {The MIT Press},
}

@article{kelly2025causal,
    title = {Sufficient, Necessary and Complete Causal Explanations in Image
             Classification},
    author = {Kelly, David A and Chockler, Hana},
    journal = {arXiv preprint arXiv:2507.23497},
    year = {2025},
}

@article{CKKS26,
    title = {Causal explanations for image classifiers},
    author = {Chockler, Hana and Kelly, David A and Kroening, Daniel and Sun,
              Youcheng},
    journal = {Journal of Artificial Intelligence Research},
    year = {2026},
}

@inproceedings{CKK25,
    title = "Multiple Different Black Box Explanations for Image Classifiers",
    author = "Hana Chockler and David Kelly and Daniel Kroening",
    note = "Publisher Copyright: {\textcopyright} 2025 The Authors.",
    year = "2025",
    doi = "10.3233/FAIA250869",
    volume = "413",
    series = "Frontiers in Artificial Intelligence and Applications",
    pages = "699 -- 706",
    booktitle = "ECAI 2025 - 28th European Conference on Artificial Intelligence
                 - Proceedings",
}

@article{CH04,
    author = {Hana Chockler and Joseph Y. Halpern},
    title = {Responsibility and Blame: {A} Structural-Model Approach},
    journal = {J. Artif. Intell. Res.},
    volume = {22},
    pages = {93--115},
    year = {2004},
}

@article{guidotti2018survey,
title={A Survey of Methods for Explaining Black Box Models},
author={Guidotti, Riccardo and Monreale, Anna and Ruggieri, Salvatore and Turini, Franco and Giannotti, Fosca and Pedreschi, Dino},
journal={ACM Computing Surveys},
volume={51},
number={5},
pages={93},
year={2018}
}

@article{verma2020counterfactual,
title={Counterfactual Explanations and Algorithmic Recourses for Machine Learning: A Review},
author={Verma, Sahaj and Dickerson, John and Hines, Keegan},
journal={arXiv preprint arXiv:2010.10596},
year={2020}
}

@inproceedings{goyal19a,
    title = {Counterfactual Visual Explanations},
    author = {Goyal, Yash and Wu, Ziyan and Ernst, Jan and Batra, Dhruv and
              Parikh, Devi and Lee, Stefan},
    booktitle = {Proceedings of the 36th International Conference on Machine
                 Learning},
    pages = {2376--2384},
    year = {2019},
    editor = {Chaudhuri, Kamalika and Salakhutdinov, Ruslan},
    volume = {97},
    series = {Proceedings of Machine Learning Research},
    month = {09--15 Jun},
    publisher = {PMLR},
}

@inproceedings{bhusalface,
    title = {FACE: Faithful Automatic Concept Extraction},
    author = {Bhusal, Dipkamal and Clifford, Michael and Rampazzi, Sara and
              Rastogi, Nidhi},
    booktitle = {The Thirty-ninth Annual Conference on Neural Information
                 Processing Systems},
    year = {2025},
}

@article{oh2022learn,
    title = {Learn-explain-reinforce: counterfactual reasoning and its guidance
             to reinforce an Alzheimer's Disease diagnosis model},
    author = {Oh, Kwanseok and Yoon, Jee Seok and Suk, Heung-Il},
    journal = {IEEE Transactions on Pattern Analysis and Machine Intelligence},
    volume = {45},
    number = {4},
    pages = {4843--4857},
    year = {2022},
    publisher = {IEEE},
}

@article{blake2024explainable,
    title = {MRxaI: Black-Box Explainability for Image Classifiers in a Medical
             Setting},
    journal = {Ceur Workshop Proceedings},
    year = {2025},
    volume = {4059},
    author = {Blake, N. and Kelly, D.A. and Pe{\~n}a, S.C. and Chanchal, A. and
              Chockler, H.},
}

@inproceedings{ribeiro2016should,
  title={" Why should i trust you?" Explaining the predictions of any classifier},
  author={Ribeiro, Marco Tulio and Singh, Sameer and Guestrin, Carlos},
  booktitle={Proceedings of the 22nd ACM SIGKDD international conference on knowledge discovery and data mining},
  pages={1135--1144},
  year={2016}
}

@article{lundberg2017unified,
  title={A unified approach to interpreting model predictions},
  author={Lundberg, Scott M and Lee, Su-In},
  journal={Advances in neural information processing systems},
  volume={30},
  year={2017}
}
